\documentclass[10pt,twocolumn,letterpaper]{article}

\usepackage{cvpr}
\usepackage{times}
\usepackage{epsfig}
\usepackage{graphicx}
\usepackage{amsmath}
\usepackage{amssymb}
\usepackage{multirow}

\usepackage{bm}
\usepackage{booktabs} % Allows the use of \toprule, \midrule and
\usepackage{epstopdf}
\usepackage{url}
\usepackage{dsfont}
\usepackage{ulem}
\usepackage{color}
\usepackage{subfigure}
\usepackage{caption}

\usepackage{algorithm,algcompatible,amssymb,amsmath}

\algnewcommand\algorithmicto{\textbf{to}}
\algnewcommand\RETURN{\State \textbf{return} }

\algnewcommand\algorithmicinput{\textbf{Input:}}
\algnewcommand\INPUT{\item[\algorithmicinput]}

\algnewcommand\algorithmicoutput{\textbf{Output:}}
\algnewcommand\OUTPUT{\item[\algorithmicoutput]}

\algnewcommand\algorithmicinitialize{\textbf{Initialize:}}
\algnewcommand\INITIALIZE{\item[\algorithmicinitialize]}

\usepackage[pagebackref=true,breaklinks=true,letterpaper=true,colorlinks,bookmarks=false]{hyperref}

\cvprfinalcopy % *** Uncomment this line for the final submission

\def\cvprPaperID{927} % *** Enter the CVPR Paper ID here

\ifcvprfinal\fi
\begin{document}

%%%%%%%%% TITLE
%\title{Dynamic Discriminative Region Erasing for Weakly Supervised Object Localization}
%\title{Complementary Discriminative Region Mining for Weakly Supervised Object Localization}
%\title{Adversarial Collaborative Modeling for Weakly Supervised Object Localization}
\title{Adversarial Complementary Learning for Weakly Supervised Object Localization}
\author{Xiaolin Zhang$^1$  \quad Yunchao Wei$^{2}$ \thanks{Corresponding Author}  \quad Jiashi Feng$^3$ \quad Yi Yang$^1$ \quad Thomas Huang$^2$\\
$^1$CAI, University of Technology Sydney \quad $^2$University of Illinois Urbana-Champaign \\
$^3$National University of Singapore \\
{\tt\small \{Xiaolin.Zhang-3@student, Yi.Yang@\}uts.edu.au \quad elefjia@nus.edu.sg}\\
{\tt\small \{yunchao,t-huang1\}@illinois.edu}
%\and
%Yunchao Wei$^2$\\
%\\
%{\tt\small wychao1987@gmail.com}
%\and
%
%\\
%{\tt\small elefjia@nus.edu.sg}
%\and
%
%\and
%
%{\tt\small }
}

\maketitle
%\thispagestyle{empty}

%%%%%%%%% ABSTRACT
\begin{abstract}
In this work, we propose \textbf{A}dversarial \textbf{Co}mplementary \textbf{L}earning (ACoL) to automatically localize integral objects of semantic interest with weak supervision. We first mathematically prove that class localization maps can be obtained by directly selecting the class-specific feature maps of the last convolutional layer, which paves a simple way to identify object regions.
We then present a simple network architecture including two parallel-classifiers for object localization. Specifically, we leverage one classification branch to dynamically localize some discriminative object regions during the forward pass. Although it is usually responsive to sparse parts of the target objects, this classifier can drive the counterpart classifier to discover new and complementary object regions by erasing its discovered regions from the feature maps. With such an adversarial learning, the two parallel-classifiers are forced to leverage complementary object regions for classification and can finally generate integral object localization together. The merits of ACoL are mainly two-fold: 1) it can be trained in an end-to-end manner; 2) dynamically erasing enables the counterpart classifier to discover complementary object regions more effectively. We demonstrate the superiority of our ACoL approach in a variety of experiments. In particular, the Top-1 localization error rate on the ILSVRC dataset is 45.14\%, which is the new state-of-the-art.

\end{abstract}

%%%%%%%%% BODY TEXT
%-------------------------------------------------------------------------
\vspace{-5mm}
\section{Introduction}
\begin{figure}
  \centering
  \includegraphics[width=0.45\textwidth]{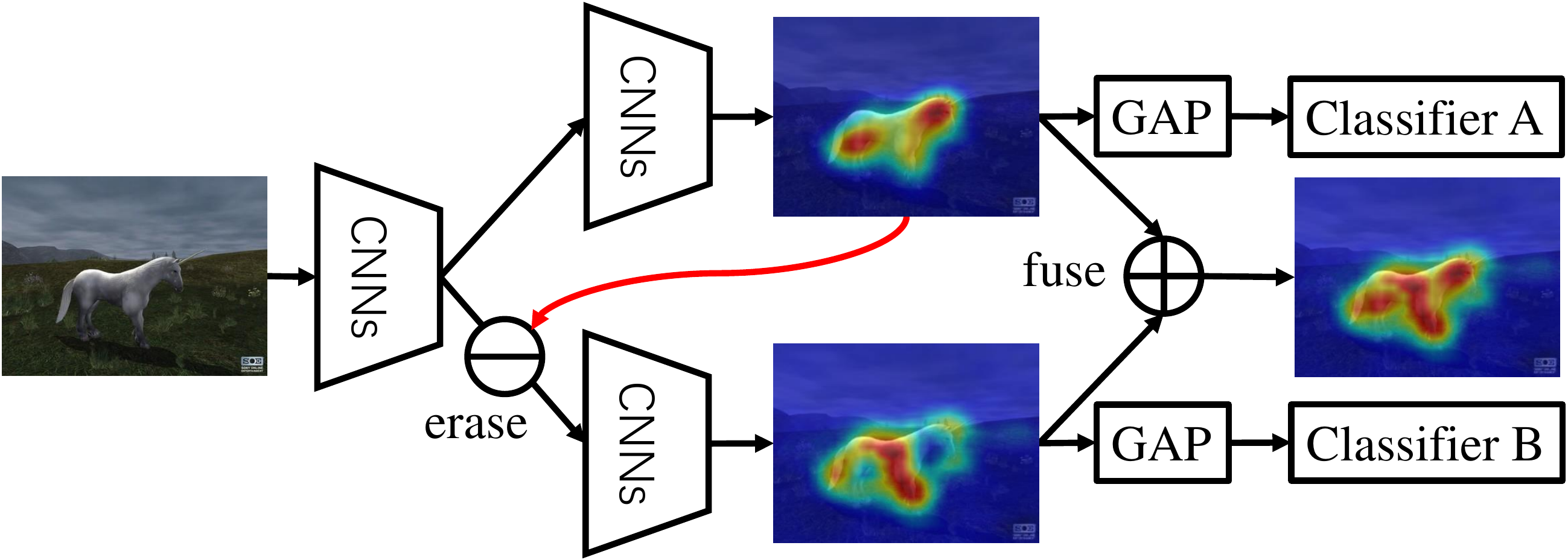}
%  \caption{Illustration of the proposed ACoL method. We prove global average pooling (GAP) can produce object localization maps during feed-forward pass. Based on this, we design the parallel adversarial classifier architecture, where complementary regions (the head and hind legs \emph{vs.} forelegs) are discovered by two classifiers (A and B) via adversarial erasing feature maps.}\label{pic-1}
  \caption{An illustration of the proposed ACoL method. We prove object localization maps can be conveniently obtained during feed-forward pass. Based on this, we design the parallel adversarial classifier architecture, where complementary regions (the head and hind legs \emph{vs.} forelegs) are discovered by two classifiers (A and B) via adversarial erasing feature maps. GAP refers to global average pooling.}\label{pic-1}
  \vspace{-5mm}
\end{figure}
Weakly Supervised Object Localization (WSOL) refers to learning object locations in a given image using the image-level labels. Currently, WSOL has drawn increasing attention since it does not require expensive bounding box annotations for training and thus can save much labour compared to fully-supervised counterparts \cite{sermanet2013overfeat, girshick2014rich, girshick15fastrcnn}.

It is a very challenging task to learn deep models for locating objects of interest using only image-level supervision.
Some pioneer works \cite{zhou2015cnnlocalization, zhang2016top} have been proposed to generate class-specific localization maps according to pre-trained convolutional classification networks.
For example, Zhou~\etal~\cite{zhou2015cnnlocalization} modified classification networks (\eg, AlexNet~\cite{krizhevsky2012imagenet} and VGG-16 \cite{simonyan2014very}) via replacing a few high-level layers by a global average pooling layer~\cite{lin2013network} and a fully connected layer, which can aggregate the features of the last convolutional layer to generate discriminative class activation maps (CAM) for the localization purpose.
However, we observe that some critical issues exist in such solutions, mainly including: 1) over-relying on category-wise discriminative features for image classification; 2) failing to localize integral regions of the target objects densely within an image.
The two issues are mainly due to the classification networks are inclined to identify patterns from the most discriminative parts for recognition, which inevitably leads to the second issue.
For instance, given an image containing a cat, the network can recognize it by identifying the head regardless of the remaining parts such as body and legs.

To tackle such issues, Wei~\etal~\cite{wei2017object} proposed an adversarial erasing (AE) approach to discover integral object regions by training additional classification networks on images whose discriminative object regions have partially been erased.
Nevertheless, one main disadvantage of AE is that it needs to train several independent classification networks for obtaining integral object regions, which costs more training time and computing resources.
Recently, Singh~\etal \cite{singh2017hide} enhanced CAM by randomly hiding the patches of input images so as to force the network to look for other discriminative parts.
However, randomly hiding patches without any high-level guidance is inefficient and cannot guarantee that networks always discover new object regions.

In this paper, we propose a novel \textbf{A}dversarial \textbf{Co}mplementary \textbf{L}earning (ACoL) approach for discovering entire objects of interest via end-to-end weakly supervised training. The key idea of ACoL is to find the complementary object regions by two adversary classifiers motivated by AE~\cite{wei2017object}.
In particular, one classifier is firstly leveraged to identify the most discriminative regions and guide the erasing operation on the intermediate feature maps. Then, we feed the erased features into its counterpart classifier for discovering new and complementary object-related regions.
Such a strategy drives the two classifiers to mine complementary object regions and finally obtain integral object localization as desired.
To easily conduct end-to-end training for ACoL, we mathematically prove that object localization maps can be obtained by directly selecting from the class-specific feature maps of the last convolutional layer, rather than using a post-inference manner in ~\cite{zhou2015cnnlocalization}. Thus discriminative object regions can be identified in a convenient way during the training forward pass according to the online inferred object localization maps.

Our approach offers multiple appealing advantages over AE~\cite{wei2017object}.
First, AE trains three networks independently for adversarial erasing.
ACoL trains two adversarial branches jointly by integrating them into a single network.
The proposed joint training framework is more capable of integrating the complementary information
among the two branches.
Second, AE adopts a recursive method to generate localization maps, and it has to forward the networks for multiple times.
Instead, our method  generates localization map by forwarding the network only once.
This advantage greatly improves the efficiency and have our method much easier for implementation.
%, thereby making our method more appropriate for real world applications.
Third, AE directly adopts CAM~\cite{zhou2015cnnlocalization} to generate localization maps.
Thus AE generates localization maps in two steps.
Differently, our method generates localization maps in one step, by selecting the feature map which best matches the groundtruth as the localization map.
We have also provided detailed proof with theoretical rigor that our method is simpler and more efficient, but yields identical results to CAM~\cite{zhou2015cnnlocalization} (see Section~\ref{CAM}).
%Fourth, our method generates localization maps at the training stage but \cite{wei2017object} generate them at the testing stage.
%The fourth advantage of our method over \cite{wei2017object} is that it is able to utilize localization map for training.
%Lastly but importantly, as we will report in \emph{A2}, our method significantly outperforms [34].

The process of ACoL is illustrated in Figure \ref{pic-1}, where an image is processed to estimate the regions of a horse. We can observe that Classifier A leverages some discriminative regions (the horse's head and hind legs) for recognition. By erasing such discriminative regions in feature maps, Classifier B is guided to use features of new and complementary object regions (the horse's forelegs) for classification. Finally, the integral target regions are obtained by fusing the object localization maps from both branches.
To validate the effectiveness of the proposed ACoL, we conduct a series of object localization experiments using the bounding boxes inferred from the generated localization maps.

To sum up, our main contributions are three-fold:
\vspace{-5pt}
\begin{itemize}
  \item We provide theoretical support of producing class-specific feature maps during the forward pass, so that object regions can be simply identified in a convenient way, which can benefit future relevant researches.
  \vspace{-5pt}
  \item We propose a novel ACoL approach to efficiently mine different discriminative regions by two adversary classifiers in a weakly supervised manner, which discover integral target regions of objects for localization.
  \vspace{-5pt}
  \item This work achieves the current state-of-the-art with the error rate of Top-1 45.14\% and Top-5 30.03\% on the ILSVRC 2016 dataset in weakly supervised setting.
\end{itemize}

%-------------------------------------------------------------------------
\section{Related Work}
\textbf{Fully supervised detection}
has been intensively studied and achieved extraordinary successes.
One of the earliest deep networks to detect objects in a one-stage manner is OverFeat~\cite{sermanet2013overfeat}, which employs a multiscale and sliding window approach to predict object boundaries. These boundaries are then applied for accumulating bounding boxes.
SSD~\cite{liu2016ssd} and YOLO~\cite{redmon2016you} use a similar one-stage method, and they are specifically designed for speeding up the detection.
Faster-RCNN~\cite{ren2015faster} utilize a novel two-stage approach and has achieved great success in the object detection.
It generates region proposals using sliding windows and predicts highly reliable object locations in a unified network in real time.
Lin~\etal~\cite{lin2017feature} presented that the performance of Faster-RCNN can be significantly improved by constructing feature pyramids with marginal extra cost.
%Most recently, Hu~\etal~\cite{hu2017relation} proposed to model the relations between objects. They propose an object relation module to process the appearance feature and geometry of objects. The model does not require additional resources and is easy to embed in existing networks.
%Girshick \etal \cite{girshick2014rich} fine-tuned a deep network supervised by ground truth bounding boxes to select the proposals generated via Selective Search \cite{uijlings2013selective}. This work was upgraded by producing proposals with regression on sliding windows to further enhance the performance~\cite{girshick15fastrcnn, ren2015faster}.
%Convolutional neural network has shown excellent abilities in image localization and detection tasks, \etc~\cite{krizhevsky2012imagenet,girshick2014rich,2014-hariharan,girshick15fastrcnn,simonyan2014very,girshick15fastrcnn,yan2013hierarchical, jiang2013salient, xia2013semantic, dai2014convolutional}.

\textbf{Weakly supervised detection and localization}
aims to apply an alternative cheaper way by only using image-level supervision~\cite{bency2016weakly,singh2017hide,bazzani2016self, wang2014video,russakovsky2015s, kim2017two, dong2017few,dong2017dual,jie2017deep,liang2015towards,oquab2015object}. Oquab \etal~\cite{oquab2015object} and Wei \etal~\cite{wei2015hcp} adopted a similar strategy to learn multi-label classification networks with max-pooling MIL. The networks are then applied to coarse object localization~\cite{oquab2015object}. Bency \etal \cite{bency2016weakly} applied a beam search method to leverage local spatial patterns, which progressively localizes bounding box candidates. Singh \etal \cite{singh2017hide} proposed a method to augment the input images by randomly hiding patches so as to look for more object regions. Similarly, Bazzani \etal \cite{bazzani2016self} analysed the scores of a classification network by randomly masking regions of input images and proposed a clustering technique to generate self-taught localization hypotheses. Deselaers \etal \cite{deselaers2012weakly} used extra images with available location annotations to learn object features and then applied a conditional random field to generally adapt the generic knowledge to specific detection tasks.
%Zhu \etal \cite{zhu2017soft} designed a network component which is nearly cost free and can be plugged into classical networks.

%However, these fully supervised training methods require expensive human-labeled annotations. To alleviate this problem, weakly supervised localization proposes to apply an alternative cheaper way by using image-level supervision~\cite{2015-dai, 2015-papandreou-weakly,wang2014video,xia2013semantic, wei2017object,russakovsky2015s}.

\textbf{Weakly supervised segmentation}
applies similar techniques to predict pixel-level labels~\cite{wei2015stc,wei2016learning,wei2017object,hou2016bottom,kolesnikov2016seed,2015-papandreou-weakly,wei2018revisiting}.
%Chaudhry~\etal~\cite{chaudhry2017discovering} propose a hierarchical approach to discover the class-agnostic attention maps for weakly segmentation.
Wei~\etal~\cite{wei2015stc} utilized extra images with simple scenes and proposed a simple to complex approach to progressively learn better pixel annotations.
Kolesnikov~\etal~\cite{kolesnikov2016seed} proposed SEC that integrates three loss functions~\ie, seeding, expansion and boundary constrain, into a unified framework to learn a segmentation network.
Wei \etal \cite{wei2017object} proposed a similar idea as ours to find more discriminative regions, they trained extra independent networks for generating class-specific activation maps with the assistance of the pre-trained networks in a post-processing step. %which wastes computational resources.

%In contrast, our proposed approach only needs to train one classification network end-to-end and it discovers discriminative regions during the forward pass in a convenient way, which benefits from the proof in Section \ref{CAM}. The proposed two classifiers can detect the integral regions of interest to improve the capability of localization.

%Although these methods contribute to solving the WSOL problem to some extent, they increase the computational expenses by using extra images or pre-processing algorithms. Recently, Zhou \etal \cite{zhou2015cnnlocalization} inspected the representative semantics of convolutional networks and proposed to generate object localization maps using a post-processing step to estimate object locations.  Singh \etal \cite{singh2017hide} proposed a method to augment the input images by randomly hiding patches so as to look for more object regions, which is inefficient and can not guarantee the network to always find useful information.
%-------------------------------------------------------------------------

\section{Adversarial Complementary Learning}

In this section, we describe details of the proposed Adversarial Complementary Learning (ACoL) approach for WSOL. We first revisit CAM \cite{zhou2015cnnlocalization} and introduce a more convenient way for producing localization maps. Then, the details of the proposed ACoL, founded on the above finding, are presented for mining high-quality object localization maps, and locating integral object regions.

\subsection{Revisiting CAM}\label{CAM}
Object localization maps have been widely used in many tasks~\cite{oquab2015object, wei2015stc, bazzani2016self, zhang2016top}, offering a promising way to visualize where deep neural networks focus on for recognition. Zhou \etal \cite{zhou2015cnnlocalization} proposed a two-step approach which can produce object localization maps by multiplying the weights from the last fully connected layer to feature maps in a classification network.

Suppose we are given a Fully Convolutional Network (FCN) with last convolutional feature maps denoted as $S \in \mathds{R}^{H \times H \times K}$, where $H \times H$ is the spatial size and $K$ is the number of channels. In \cite{zhou2015cnnlocalization}, the feature maps are fed into a Global Average Pooling (GAP) \cite{lin2013network} layer followed by a fully connected layer. A softmax layer is applied on the top for classification. We denote the average value of the $k_{th}$ feature map as $s_k = \frac{\sum_{i,j} (S_k)_{i,j}}{H \times H}, k=0,1,...,K-1$, where $(S_k)_{i,j}$ is the element of the $k_{th}$ feature map $S_k$ at the $i_{th}$ row and the $j_{th}$ column. The weight matrix of the fully connected layer is denoted as $W^{fc} \in \mathds{R}^{K \times C}$, where $C$ is the number of target classes. Here, we ignore the bias term for convenience. Therefore, for the target class $c$, the input of the $c_{th}$ softmax node $y^{fc}_c$ can be defined as
\begin{equation}\label{eq1}
\small
y^{fc}_c = \sum_{k=0}^{K-1}s_k W^{fc}_{k,c},
\vspace{-5pt}
\end{equation}
where $W^{fc}_{k, c} \in \mathds{R}$ denotes the element of the matrix $W^{fc}$ at the $k_{th}$ row and the $c_{th}$ column. The row $W^{fc}_{k,c}, k=0,1,...,K-1$ contributes to calculating the value $y^{fc}_c$. Therefore, the object localization map $A^{fc}_c$ of class $c$ proposed in \cite{zhou2015cnnlocalization} can be obtained by aggregating the feature map $S$ as follows, %Eq. \eqref{eq2}
\begin{equation}\label{eq2}
\small
  A^{fc}_c = \sum_{k=0}^{K-1} S_{k} \cdot W^{fc}_{k,c}.
  \vspace{-5pt}
\end{equation}

CAM provides a useful way to inspect and locate the target object regions,
but it needs an extra step to generate object localization maps after the forward pass. In this work, we reveal that object localization maps can be conveniently obtained by directly selecting from the feature maps of the last convolutional layer. Recently, some methods \cite{hwang2016self, chaudhry2017discovering} have already obtained localization maps like this, but we are the first to prove this convenient approach can generate same-quality localization maps with CAM, which is meaningful and contributes to embedding localization maps into complex networks.
\begin{figure}[t]
  %input
  \centering
  \begin{minipage}[c]{0.450\textwidth}
    \begin{minipage}[c]{0.04\textwidth}
      \raggedright
      \vspace{-34pt}
      \caption*{\rotatebox{90}{Image}}
      \end{minipage}
      \begin{minipage}[t]{1.0\textwidth}
      \includegraphics[width=0.227\textwidth,height=0.227\textwidth]{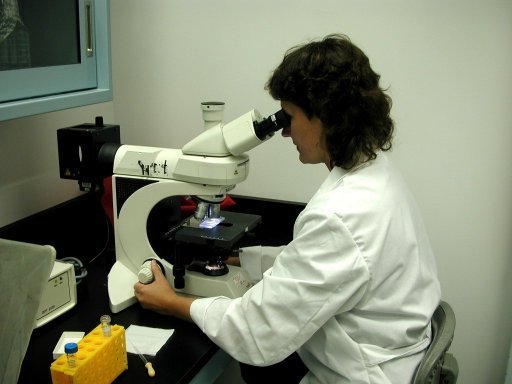}
      \includegraphics[width=0.227\textwidth,height=0.227\textwidth]{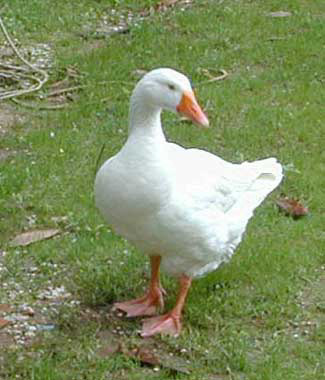}
      \includegraphics[width=0.227\textwidth,height=0.227\textwidth]{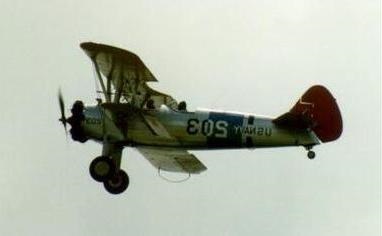}
      \includegraphics[width=0.227\textwidth,height=0.227\textwidth]{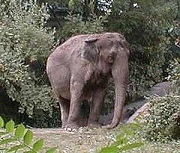}
    \end{minipage}
  \end{minipage}
  \begin{minipage}[c]{0.450\textwidth}
     \begin{minipage}[c]{0.04\textwidth}
      \raggedright
      \vspace{-34pt}
      \caption*{\rotatebox{90}{CAM}}
      \end{minipage}
      \begin{minipage}[t]{1.0\textwidth}
          \includegraphics[width=0.227\textwidth,height=0.227\textwidth]{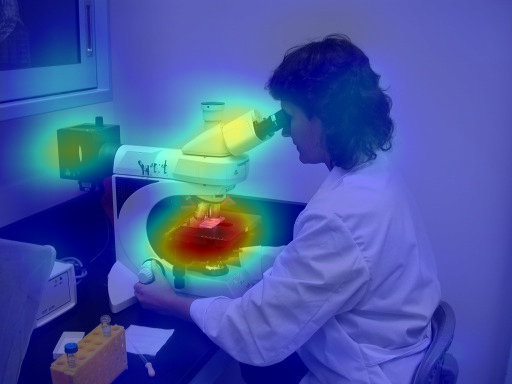}
          \includegraphics[width=0.227\textwidth,height=0.227\textwidth]{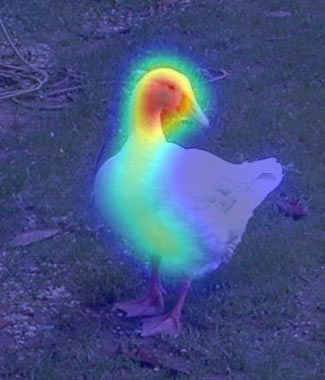}
          \includegraphics[width=0.227\textwidth,height=0.227\textwidth]{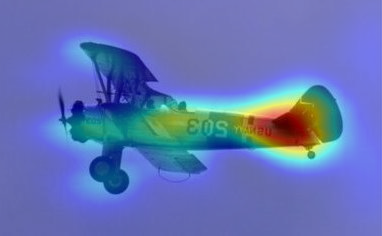}
          \includegraphics[width=0.227\textwidth,height=0.227\textwidth]{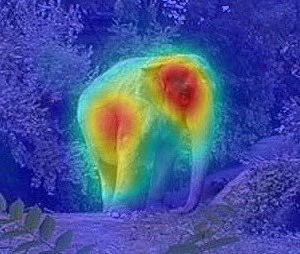}
     \end{minipage}
  \end{minipage}
  \begin{minipage}[c]{0.450\textwidth}
     \begin{minipage}[c]{0.04\textwidth}
      \raggedright
      \vspace{-34pt}
      \caption*{\rotatebox{90}{Ours}}
      \end{minipage}
      \begin{minipage}[t]{1.0\textwidth}
          \includegraphics[width=0.227\textwidth,height=0.227\textwidth]{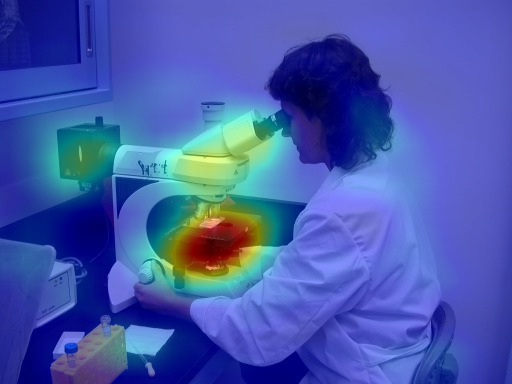}
          \includegraphics[width=0.227\textwidth,height=0.227\textwidth]{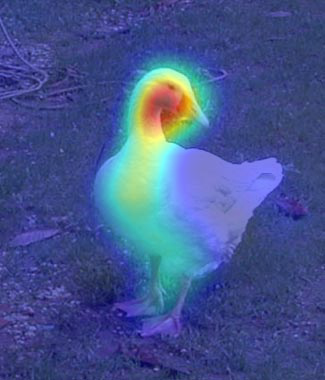}
          \includegraphics[width=0.227\textwidth,height=0.227\textwidth]{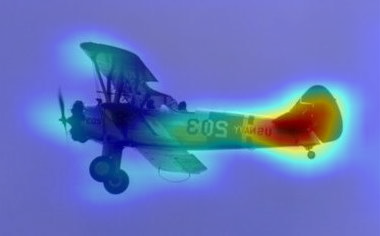}
          \includegraphics[width=0.227\textwidth,height=0.227\textwidth]{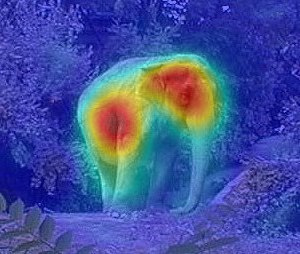}
     \end{minipage}
  \end{minipage}
  \caption{Comparison of methods for generating localization maps. Our method can produce the same-quality maps as CAM~\cite{zhou2015cnnlocalization} but in a more convenient way.}\label{fig3}
  \vspace{-4mm}
\end{figure}
\begin{figure*}[t]
  \centering
  \includegraphics[width=0.75\textwidth]{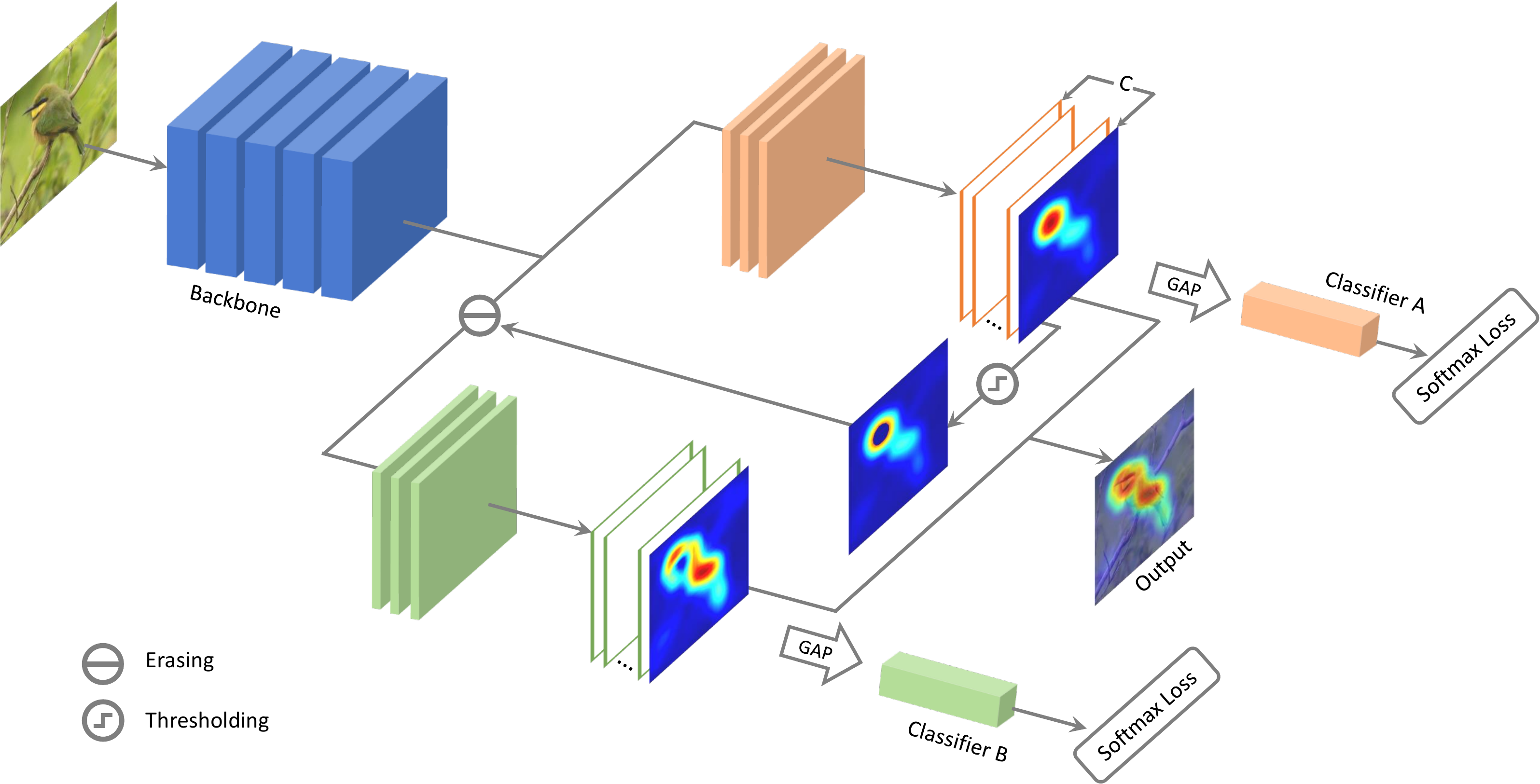}
  \caption{Overview of the proposed ACoL approach. The input images are processed by Backbone to extract mid-level feature maps, which are then fed into two parallel-classifiers for discovering complementary object regions. Each classifier consists of several convolutional layers followed by a global average pooling (GAP) layer and a softmax layer. Different from Classifier A, the input feature maps of Classifier B are erased with the guidance of the object localization maps from Classifier A. Finally, the object maps from the two classifiers are fused for localization.}\label{pic-2}
  \vspace{-5mm}
\end{figure*}
In the following, we provide both theoretical proof and visualized comparison to support our discovery. Given the output feature maps $S$ of an FCN, we add a convolutional layer of $C$ channels with the kernel size of $1 \times 1$, stride 1 on top of the feature maps $S$. Then, the output is fed into a GAP layer followed by a softmax layer for classification. Suppose the weight matrix of the $1 \times 1$ convolutional layer is $W^{conv} \in \mathds{R}^{K \times C}$. We define the localization maps $A^{conv}_c, c=0,1,...,C-1$ as the output feature maps of the $1 \times 1$ convolutional layer and $A^{conv}_c$ can be calulated by
\begin{equation}\label{eq3}
  A^{conv}_c = \sum_{k=0}^{K-1} S_{k} \cdot W^{conv}_{k,c},
  \vspace{-5pt}
\end{equation}
where $W^{conv}_{k, c} \in \mathds{R}$ denotes the element of the matrix $W^{conv}$ at the $k_{th}$ row and the $c_{th}$ column. Therefore, the $c_{th}$ input value $y^{conv}_c$ of the softmax layer is the average value of $A^{conv}_c$. So, $y^{conv}_c$ can be calculated by
\begin{equation}\label{eq4}
  y^{conv}_c = \frac{\sum_{i,j} (A^{conv}_{c})_{i,j}}{H \times H}.
  \vspace{-5pt}
\end{equation}

It is observed that the $y^{fc}_c$ and $y^{conv}_c$ are equal if we initialize the parameters of the both networks in the same way. Also, $A^{fc}_c$ and $A^{conv}_c$ have the same mathematical form. Therefore, we get the same-quality object localization maps $A^{fc}_c$ and $A^{conv}_c$ after the networks are convergent. In practice, the object localization maps from both methods are very similar and highlight the same target regions expect for some marginal differences caused by the stochastic optimization process. Figure \ref{fig3} compares the object localization maps generated by CAM and our revised approach. We observe that the both approaches can generate the same quality maps and highlight the same regions in a given image. However, with our revised method, the object localization maps can be directly obtained in the forward pass rather than a post-processing step proposed in CAM.
%, but our method is much more convenient for generating localization maps.
\subsection{The proposed ACoL}
The mathematical proof in Section~\ref{CAM} provides theoretical support of the proposed ACoL. We identify that deep classification networks usually leverage the unique pattern of a specific category for recognition and the generated object localization maps can only highlight a small region of the target object instead of the entire object. Our proposed ACoL aims at discovering the integral object regions through an adversarial learning manner. In particular, it includes two classifiers, which can mine different but complementary regions of the target object in a given image.

Figure \ref{pic-2} shows the architecture of the proposed ACoL, including three components, Backbone, Classifier A and Classifier B. Backbone is a fully convolutional network acting as a feature extractor, which takes the original RGB images as input and produces high-level position-aware feature maps of multiply channels. The feature maps from Backbone are then fed into the following parallel classification branches. The object localization maps for each classifier can be conveniently obtained as described in Section~\ref{CAM}. Both branches consist of the same number of convolutional layers followed by a GAP layer and a softmax layer for classification. The input feature maps of the two classifiers are different. In particular, the input features of Classifier B are erased with the guidance of the mined discriminative regions produced by Classifier A. We identify the discriminative regions by conducting a threshold on the localization maps of Classifier A. The corresponding regions within the input feature maps for Classifier B are then erased in an adversarial manner via replacing the values by zeros. Such an operation encourage Classifier B to leverage features from other regions of the target object for supporting image-level labels. Finally, the integral localization map of the target object will be obtained by combining the localization maps produced by the two branches.

%We now explain the ACoL process more formally.
Formally, we denote the training image set as $I=\{(I_i, y_i)\}^{N-1}_{i=0}$, where $y_i$ is the label of the image $I_i$ and $N$ is the number of images. The input image $I_i$ is firstly transformed by Backbone $f(\theta_0)$ to the spatial feature maps $S \in \mathds{R}^{H_1 \times H_1 \times K}$ with $K$ channels and $H_1 \times H_1$ resolution. We use $\theta$ to denote the learnable parameters of the CNN. Classifier A is denoted as $f(\theta_{A})$ which can generate object map $M^A \in \mathds{R}^{H_2 \times H_2}$ of the size $H_2 \times H_2$ given the input feature maps $S$ in a weakly supervised manner, as explained in Section~\ref{CAM}. $M^A$ usually highlights the unique discriminative regions for the target class.

We identify the most discriminative region as the set of pixels whose value is larger than the given threshold $\delta$ in object localization maps. $M^A$ is resized by linear interpolation to $H_1 \times H_1$ if $H_1 \neq H_2$. We erase the discriminative regions in $S$ according to the mined discriminative regions. Let $\tilde{S}$ denote the erased feature maps, which can be generated via replacing the pixel values of the identified discriminative regions by zeros. Classifier B $f(\theta_{B})$ can generate the object localization maps $M^B \in \mathds{R}^{H_2 \times H_2}$ with the input $\tilde{S}$. Then, the parameters $\theta$ of the network can be updated by back-propagation. Finally, we can obtain the integral object map for the class $c$ by merging the two maps $M^A$ and $M^B$. Concretely, we normalize both maps to the range $[0,1]$ and denote them as $\bar{M}^A$ and $\bar{M}^B$ . The fused object localization map $\bar{M}^{fuse}$ is calculated by $\bar{M}^{fuse}_{i,j}=\max(\bar{M}^A_{i,j},\bar{M}^B_{i,j})$, where $\bar{M}_{i,j}$ is the element of the normalized map $\bar{M}$ at the $i_{th}$ row and $j_{th}$ column. The whole process is trained in an end-to-end way. Both classifiers adopt the cross entropy loss function for training.
Algorithm \ref{alg1} illustrates the training procedure of the proposed ACoL approach.

\begin{algorithm}
\small
\caption{Training algorithm for ACoL}
\label{alg1}
\begin{algorithmic}[1]
\INPUT Training data $\emph{I}=\{(I_i, y_i)\}_{i=1}^N$, threshold $\delta$
\WHILE {training is not convergent}
\STATE Update feature maps $S \leftarrow f(\theta_0, I_i)$
\STATE Extract localization map $M^A \leftarrow f(\theta_A, S, y_i)$
\STATE Discover the discriminative region $R = \bar{M}^A > \delta$
\STATE Obtain erased feature maps $\tilde{S} \leftarrow erase(S,R)$ %$\tilde{S} \leftarrow S \backslash R$
\STATE Extract localization map $M^B \leftarrow f(\theta_A, S, y_i)$
\STATE Obtain fused map $\bar{M}^{fuse}_{i,j}=\max(\bar{M}^A_{i,j},\bar{M}^A_{i,j})$
\STATE Update $\theta_0, \theta_A$ and $\theta_B$
\ENDWHILE
\OUTPUT $\bar{M}^{fuse}$
\end{algorithmic}
\end{algorithm}
\vspace{-2mm}

During testing, we extract the fused object maps according to the predicted class and resize them to the same size with the original images by linear interpolation. For fair comparison, we apply the same strategy detailed in \cite{zhou2015cnnlocalization} to produce object bounding boxes based on the generated object localization maps. In particular, we firstly segment the foreground and background by a fixed threshold. Then, we seek the tight bounding boxes covering the largest connected area in the foreground pixels. For more details please refer to \cite{zhou2015cnnlocalization}.

%-------------------------------------------------------------------------
\section{Experiments}

\subsection{Experiment setup}
\textbf{Datasets and evaluation metrics}
We evaluate the classification and localization accuracy of ACoL on two datasets, \ie, ILSVRC 2016 \cite{2009-imagenet, ILSVRC15} and CUB-200-2011~\cite{WahCUB_200_2011}. ILSVRC 2016 contains 1.2 million images of 1,000 categories for training. We compare our approach with other approaches on the \emph{validation} set which has 50,000 images. CUB-200-2011~\cite{WahCUB_200_2011} has 11,788 images of 200 categories with 5,994 images for training and 5,794 for testing. We leverage the localization metric suggested by~\cite{ILSVRC15} for comparison. The metric calculates the percentage of the images whose bounding boxes have over 50\% IoU with the ground-truth. In addition, we also implement our approach on Caltech-256~\cite{griffin2007caltech} to visualize the outstanding performance in locating the integral target object.

\textbf{Implementation details}
We evaluate the proposed ACoL using VGGnet \cite{simonyan2014very} and GoogLeNet \cite{szegedy2014going}. Particularly, we remove the layers after \emph{conv5-3} (from \emph{pool5} to \emph{prob}) of VGG-16 network and the last \emph{inception} block of GoogLeNet.
Then, we add two convolutional layers of kernel size $3 \times 3$, stride 1, pad 1 with 1024 units and a convolutional layer of size $1 \times 1$, stride 1 with 1000 units (200 and 256 units for CUB-200-2011 and Caltech-256 datasets, respectively).
As the proof in Section \ref{CAM}, localization maps can be conveniently obtained from the feature maps of the $1 \times 1$ convolutional layer. Finally, a GAP layer and a softmax layer are added on the top of the convolutional layers. Both networks are fine-tuned on the pre-trained weights of ILSVRC \cite{ILSVRC15}. The input images are randomly cropped to $224 \times 224$ pixels after being resized to $256 \times 256$ pixels. We test different erasing thresholds $\delta$ from 0.5 to 0.9. In testing, the threshold $\delta$ maintains constant \wrt the value in training. For classification results, we average the scores from the softmax layer with 10 crops (4 corners plus center, same with horizontal flip).  We train the networks on NVIDIA GeForce TITAN X GPU with 12GB memory.

\subsection{Comparisons with the state-of-the-arts}
%We conduct extensive experiments to evaluate the proposed method in both classification and localization tasks.
%We compare it with the state-of-the-art methods on ILSVRC validation set and CUB-200-2011 test set.
\begin{table}\setlength{\tabcolsep}{13pt}
  \centering
  \small
\begin{tabular}{l|c|c}
  \hline
  \hline
  Methods & top-1 err. & top-5 err. \\
  \hline
  GoogLeNet-GAP \cite{zhou2015cnnlocalization} & 35.0 & 13.2 \\
  GoogLeNet & 30.6 & 10.5 \\
  GoogLeNet-ACoL(Ours) & 29.0 & 11.8 \\
  \hline
  VGGnet-GAP \cite{zhou2015cnnlocalization} & 33.4 & 12.2 \\
  VGGnet  & 31.2 & 11.4 \\
  VGGnet-ACoL(Ours) & 32.5 & 12.0 \\
  \hline
  \hline
\end{tabular}
  \caption{Classification error on ILSVRC validation set.}\label{tab1}
  \vspace{-5mm}
\end{table}
\begin{table}\setlength{\tabcolsep}{5pt}
  \centering
  \small
  \begin{tabular}{l|c|c}
    \hline
    \hline
    Methods & Train/Test anno. & err. \\
    \hline
    Alignments \cite{gavves2015local} & w/\/o & 46.4 \\
    Alignments \cite{gavves2015local} & BBox & 33.0 \\
    DPD \cite{zhang2013deformable} & BBox+Parts & 49.0 \\
    DeCAF+DPD \cite{donahue2014decaf} & BBox+Parts & 35.0 \\
    PANDA R-CNN \cite{zhang2014part} & BBox+Parts & 23.6  \\
    \hline
    GoogLeNet-GAP on full image \cite{zhou2015cnnlocalization} & w/\/o &37.0 \\
    GoogLeNet-GAP on crop \cite{zhou2015cnnlocalization} & w/\/o & 32.2 \\
    GoogLeNet-GAP on BBox \cite{zhou2015cnnlocalization} & BBox & 29.5 \\
    \hline
    VGGnet-ACoL(Ours) & w/\/o & 28.1 \\
    \hline
    \hline
  \end{tabular}
  \caption{Classification error on fine-grained CUB-200-2011 test set.}\label{tab2}
  \vspace{-2mm}
\end{table}

\textbf{Classification:}~Table \ref{tab1} shows the Top-1 and Top-5 error on the ILSVRC validation set.
Our proposed methods GoogLeNet-ACoL and VGGnet-ACoL achieve sightly better classification results than GoogLeNet-GAP and VGGnet-GAP, respectively, and are comparable to the original GoogLeNet and VGGnet.
For the fine-grained recognition dataset CUB-200-2011, it also achieves remarkable performance. Table \ref{tab2} summarizes the benchmark approaches for classification with or without (w/\/o) bounding box annotations. We find our VGGnet-ACoL achieves the lowest error 28.1\% among all the methods without using bounding box.
%Also, it outperforms DPD \cite{zhang2013deformable}, DeCAF+DPD \cite{donahue2014decaf} and GoogLeNet-GAP \cite{zhou2015cnnlocalization} which utilize the bounding box annotation in training for classification.

To summarize, the proposed method can enable the networks to achieve equivalent classification performance with the original networks though our modified networks actually do not use fully connected layers.
We attribute it to the erasing operation which guides the network to discover more discriminative patterns so as to obtain better classification performance.

\begin{table}\setlength{\tabcolsep}{10pt}
  \centering
  \small
  \begin{tabular}{l|c|c}
    \hline
    \hline
     Methods & top-1 err. & top-5 err. \\
    \hline
     Backprop on GoogLeNet \cite{simonyan2013deep} & 61.31 & 50.55 \\
     GoogLeNet-GAP \cite{zhou2015cnnlocalization} & 56.40 & 43.00 \\
     GoogLeNet-HaS-32 \cite{singh2017hide} & 54.53 & - \\
     GoogLeNet-ACoL(Ours) & 53.28 & 42.58 \\
     GoogLeNet-ACoL*(Ours) & 53.28 & 35.22 \\
    \hline
     Backprop on VGGnet \cite{simonyan2013deep} & 61.12 & 51.46 \\
     VGGnet-GAP \cite{zhou2015cnnlocalization} & 57.20 & 45.14 \\
     VGGnet-ACoL(Ours) & 54.17 & 40.57 \\
     VGGnet-ACoL*(Ours) & 54.17 & 36.66 \\
    \hline
    \hline
  \end{tabular}
  \caption{Localization error on ILSVRC validation set (* indicates methods which improve the Top-5 performance only using predictions with high scores).}\label{tab3}
  \vspace{-3mm}
\end{table}
\begin{table}\setlength{\tabcolsep}{14pt}
  \centering
  \small
  \begin{tabular}{l|c|c}
    \hline
    \hline
    Methods & top-1 err. & top-5 err. \\
    \hline
    GoogLeNet-GAP \cite{zhou2015cnnlocalization} & 59.00 & - \\
    VGGnet-ACoL(Ours) & 54.08 & 43.49 \\
    VGGnet-ACoL*(Ours) & 54.08 & 39.05  \\
    \hline
    \hline
  \end{tabular}
  \caption{Localization error on CUB-200-2011 test set.}\label{tab4}
  \vspace{-5mm}
\end{table}
\begin{table}\setlength{\tabcolsep}{6pt}
  \centering
  \small
  \begin{tabular}{l|c|c}
    \hline
    \hline
    Methods & top-1 err. & top-5 err. \\
%    \hline
%    VGGnet-ACoL-GoogLeNet & 52.90/30.65 & 41.72/10.49 \\
    \hline
    VGGnet-ACoL-ResNet-50 & 49.82/26.22 & 40.38/8.47 \\
    VGGnet-ACoL-ResNet-101 & 49.26/24.90 & 40.08/7.80 \\
    VGGnet-ACoL-ResNet-152 & 48.96/24.39 & 39.97/7.59 \\
    \hline
    VGGnet-ACoL-DPN-92  & 46.30/17.70 & 38.96/3.83 \\
    VGGnet-ACoL-DPN-98  & 46.16/17.42 & 38.99/3.67 \\
    VGGnet-ACoL-DPN-131  & 46.06/17.08 & 38.85/3.42 \\
    VGGnet-ACoL-DPN-ensemble  & 45.14/15.47 & 38.45/2.70 \\
    VGGnet-ACoL-DPN-ensemble*  & 45.14/15.47 & 30.03/2.70 \\
    \hline
    \hline
  \end{tabular}
  \caption{Localization/Classification error on ILSVRC validation set with the state-of-the-art classification results.}\label{tab5}
  \vspace{-5mm}
\end{table}
\begin{figure*}[t]
  \small
  \begin{minipage}[t]{0.5\textwidth}
  \begin{subfigure}[CUB-200-2011]{
  \begin{minipage}[c]{1.0\textwidth}
  %input
  \centering
  \begin{minipage}[c]{1.0\textwidth}
      \begin{minipage}[c]{0.04\textwidth}
          \raggedright
          \vspace{-34pt}
          \caption*{\rotatebox{90}{Image}}
      \end{minipage}
      \begin{minipage}[t]{1.0\textwidth}
         \includegraphics[width=0.2277\textwidth,height=0.2277\textwidth]{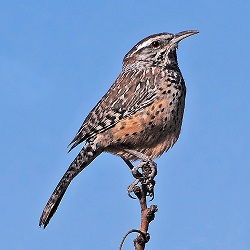}
         \includegraphics[width=0.227\textwidth,height=0.227\textwidth]{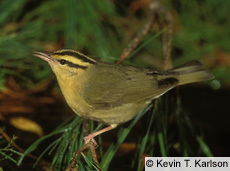}
         \includegraphics[width=0.227\textwidth,height=0.227\textwidth]{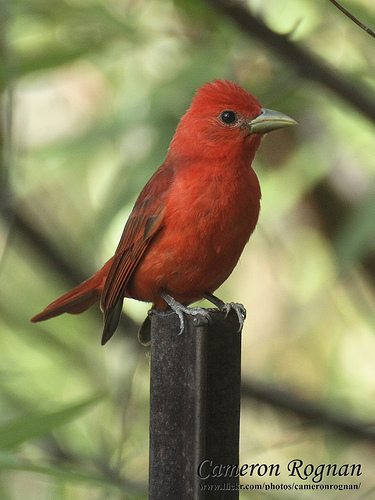}
         \includegraphics[width=0.227\textwidth,height=0.227\textwidth]{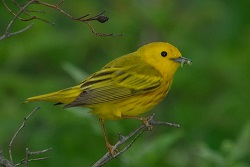}
    \end{minipage}
  \end{minipage}
  \begin{minipage}[c]{1.0\textwidth}
     \begin{minipage}[c]{0.04\textwidth}
      \raggedright
      \vspace{-34pt}
      \caption*{\rotatebox{90}{CAM}}
      \end{minipage}
      \begin{minipage}[t]{1.0\textwidth}
         \includegraphics[width=0.227\textwidth,height=0.227\textwidth]{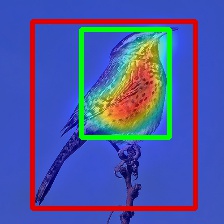}
         \includegraphics[width=0.227\textwidth,height=0.227\textwidth]{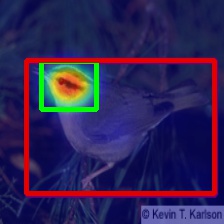}
         \includegraphics[width=0.227\textwidth,height=0.227\textwidth]{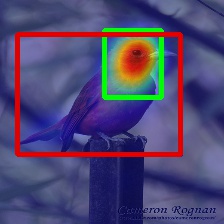}
         \includegraphics[width=0.227\textwidth,height=0.227\textwidth]{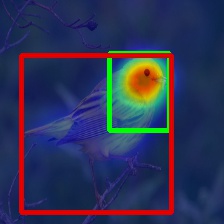}
     \end{minipage}
  \end{minipage}
  \begin{minipage}[c]{1.0\textwidth}
     \begin{minipage}[c]{0.04\textwidth}
      \raggedright
      \vspace{-34pt}
      \caption*{\rotatebox{90}{Ours}}
      \end{minipage}
      \begin{minipage}[t]{1.0\textwidth}
         \includegraphics[width=0.227\textwidth,height=0.227\textwidth]{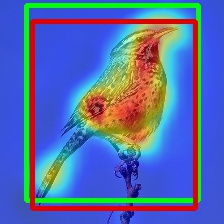}
         \includegraphics[width=0.227\textwidth,height=0.227\textwidth]{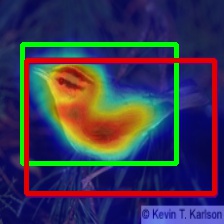}
         \includegraphics[width=0.227\textwidth,height=0.227\textwidth]{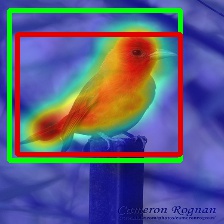}
         \includegraphics[width=0.227\textwidth,height=0.227\textwidth]{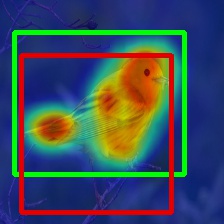}

      \end{minipage}
  \end{minipage}
  \end{minipage}
  }
  \end{subfigure}
  \end{minipage}
  \begin{minipage}[t]{0.5\textwidth}
  \begin{subfigure}[ILSVRC]{
  \begin{minipage}[t]{1.0\textwidth}
  \includegraphics[width=0.227\textwidth,height=0.227\textwidth]{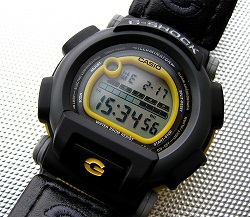}
  \includegraphics[width=0.227\textwidth,height=0.227\textwidth]{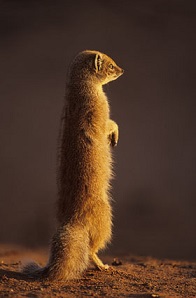}
  \includegraphics[width=0.227\textwidth,height=0.227\textwidth]{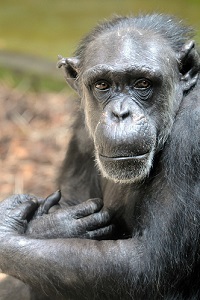}
  \includegraphics[width=0.227\textwidth,height=0.227\textwidth]{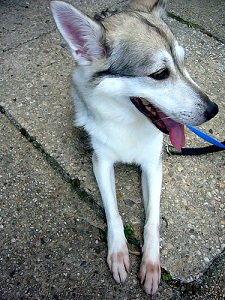}

  \includegraphics[width=0.227\textwidth,height=0.227\textwidth]{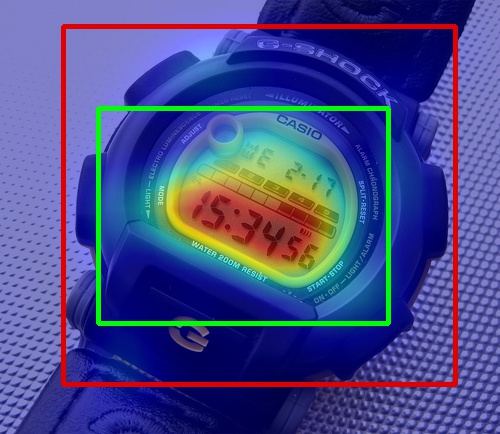}
  \includegraphics[width=0.227\textwidth,height=0.227\textwidth]{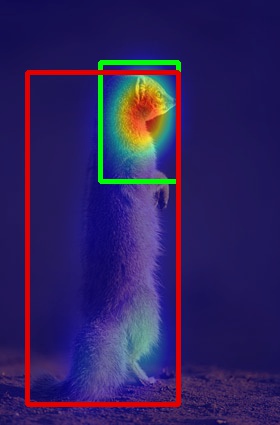}
  \includegraphics[width=0.227\textwidth,height=0.227\textwidth]{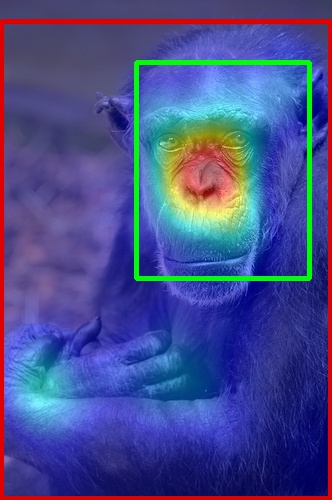}
  \includegraphics[width=0.227\textwidth,height=0.227\textwidth]{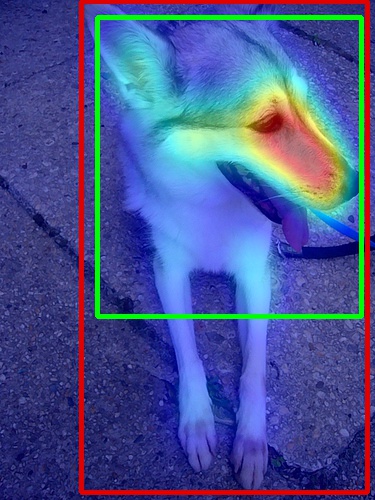}

  \includegraphics[width=0.227\textwidth,height=0.227\textwidth]{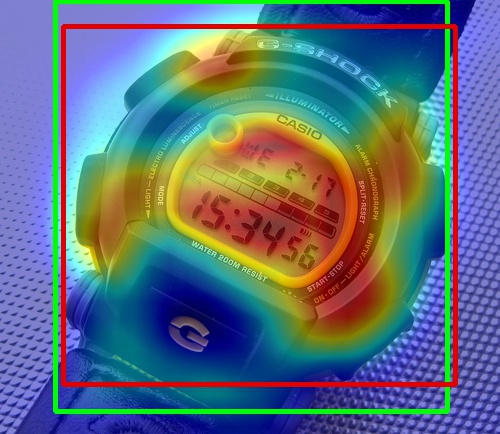}
  \includegraphics[width=0.227\textwidth,height=0.227\textwidth]{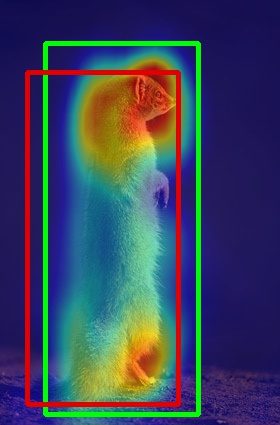}
  \includegraphics[width=0.227\textwidth,height=0.227\textwidth]{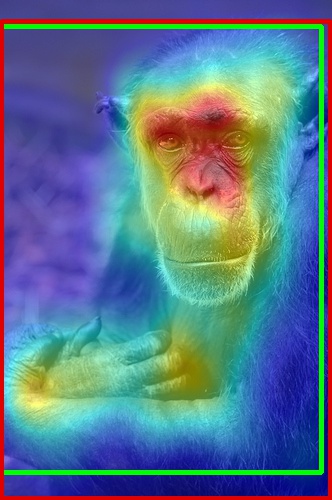}
  \includegraphics[width=0.227\textwidth,height=0.227\textwidth]{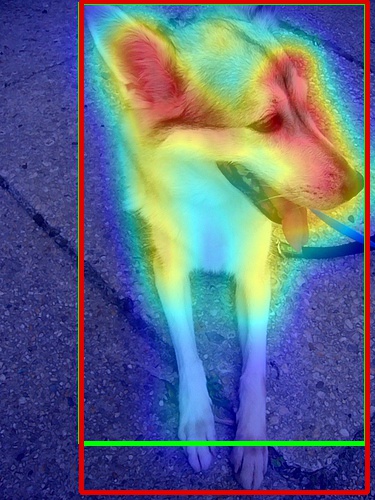}
  \end{minipage}
  }
  \end{subfigure}
  \end{minipage}

  \caption{Comparison with CAM method. Our method can locate larger object regions to improve localization performance (ground-truth bounding boxes are in red and the predicted are in green).}\label{bbox}
  \vspace{-4mm}
\end{figure*} 
\textbf{Localization:}~Table \ref{tab3} illustrates the localization error on the ILSVRC \emph{val} set. We observe that our ACoL approach outperforms all baselines. VGGnet-ACoL is significantly better than VGGnet-GAP and GoogLeNet-ACoL also achieves better performance than GoogLeNet-HaS-32 which adopts the strategy of randomly erasing the input images. We illustrate the localization performance on the CUB-200-2011 dataset in Table \ref{tab4}. Our method outperforms GoogLeNet-GAP by 4.92\% in Top-1 error.

We further improve the localization performance by combining our localization results with the state-of-the-art classification results, \ie, ResNet~\cite{he2016deep} and DPN~\cite{chen2017dual}, to break the limitation of classification when calculating localization accuracy. As shown in Table \ref{tab5}, the localization accuracy constantly improves with the classification results getting better.  We have a boost to 45.14\% in Top-1 error and 38.45\% in Top-5 error when applying the classification results generated from the ensemble DPN. In addition, we boost the Top-5 localization performance (indicated by *) by only selecting the bounding boxes from the top three predicted classes following \cite{zhou2015cnnlocalization} and VGGnet-ACoL-DPN-ensemble* achieves 30.03\% on ILSVRC.

\begin{figure*}
  \begin{subfigure}[ILSVRC]{
  \begin{minipage}[t]{0.115\textwidth}
  \setlength{\abovecaptionskip}{0pt}
  \caption*{Image}
  \includegraphics[width=1.0\textwidth,height=1.0\textwidth]{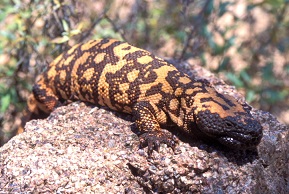}
  \includegraphics[width=1.0\textwidth,height=1.0\textwidth]{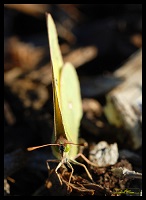}
  \includegraphics[width=1.0\textwidth,height=1.0\textwidth]{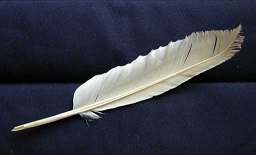}
  \includegraphics[width=1.0\textwidth,height=1.0\textwidth]{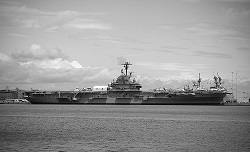}
  \end{minipage}
  \begin{minipage}[t]{0.115\textwidth}
    \setlength{\abovecaptionskip}{0pt}
    \caption*{Classifier A}
    \includegraphics[width=1\textwidth,height=1\textwidth]{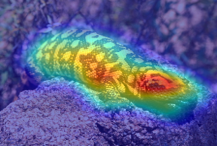}
    \includegraphics[width=1.0\textwidth,height=1.0\textwidth]{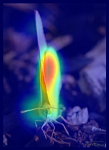}
    \includegraphics[width=1.0\textwidth,height=1.0\textwidth]{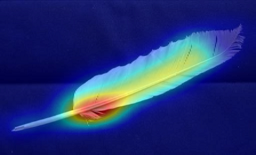}
    \includegraphics[width=1.0\textwidth,height=1.0\textwidth]{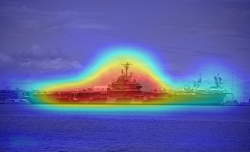}
  \end{minipage}
  \begin{minipage}[t]{0.115\textwidth}
    \setlength{\abovecaptionskip}{0pt}
  \caption*{Classifier B}
    \includegraphics[width=1\textwidth,height=1\textwidth]{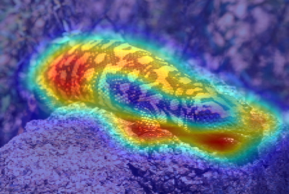}
    \includegraphics[width=1.0\textwidth,height=1.0\textwidth]{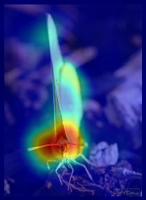}
    \includegraphics[width=1.0\textwidth,height=1.0\textwidth]{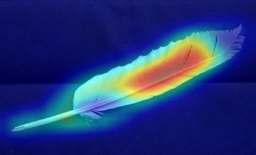}
    \includegraphics[width=1.0\textwidth,height=1.0\textwidth]{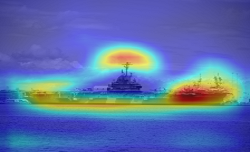}
  \end{minipage}
  \begin{minipage}[t]{0.115\textwidth}
    \setlength{\abovecaptionskip}{0pt}
  \caption*{Fusion}
    \includegraphics[width=1\textwidth,height=1\textwidth]{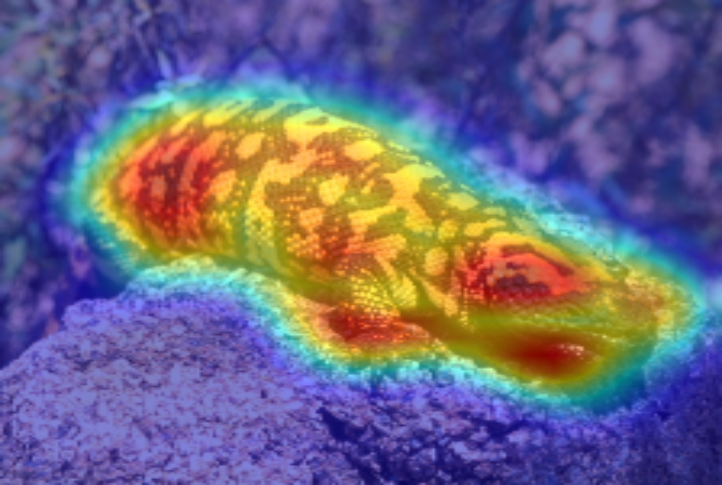}
    \includegraphics[width=1.0\textwidth,height=1.0\textwidth]{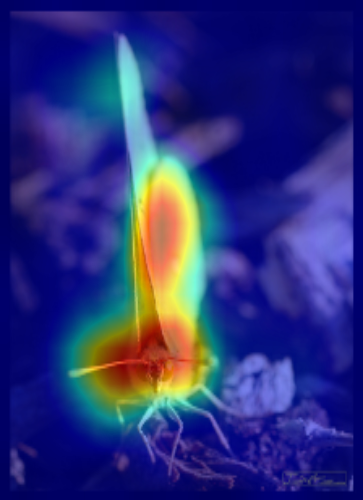}
    \includegraphics[width=1.0\textwidth,height=1.0\textwidth]{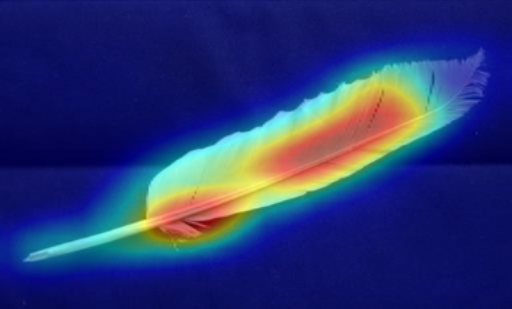}
    \includegraphics[width=1.0\textwidth,height=1.0\textwidth]{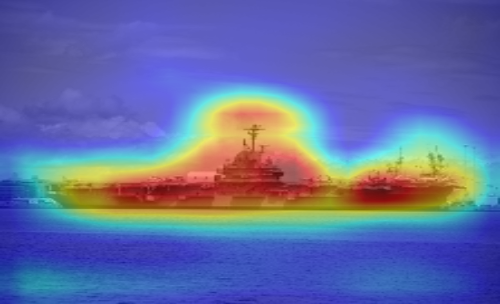}
  \end{minipage}
  \hspace{16pt}
  \begin{minipage}[t]{0.115\textwidth}
    \setlength{\abovecaptionskip}{0pt}
    \caption*{Image}
    \includegraphics[width=1.0\textwidth,height=1.0\textwidth]{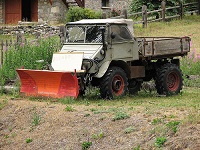}
    \includegraphics[width=1.0\textwidth,height=1.0\textwidth]{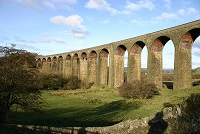}
    \includegraphics[width=1.0\textwidth,height=1.0\textwidth]{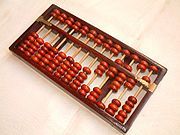}
    \includegraphics[width=1.0\textwidth,height=1.0\textwidth]{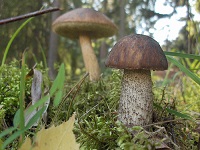}
  \end{minipage}
  \begin{minipage}[t]{0.115\textwidth}
    \setlength{\abovecaptionskip}{0pt}
    \caption*{Classifier A}
    \includegraphics[width=1.0\textwidth,height=1.0\textwidth]{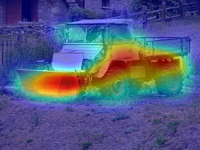}
    \includegraphics[width=1.0\textwidth,height=1.0\textwidth]{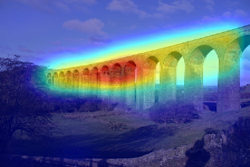}
    \includegraphics[width=1.0\textwidth,height=1.0\textwidth]{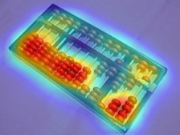}
    \includegraphics[width=1.0\textwidth,height=1.0\textwidth]{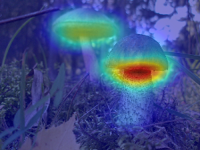}
  \end{minipage}
  \begin{minipage}[t]{0.115\textwidth}
    \setlength{\abovecaptionskip}{0pt}
    \caption*{Classifier B}
    \includegraphics[width=1.0\textwidth,height=1.0\textwidth]{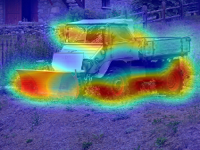}
    \includegraphics[width=1.0\textwidth,height=1.0\textwidth]{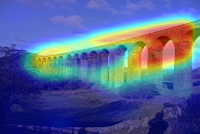}
    \includegraphics[width=1.0\textwidth,height=1.0\textwidth]{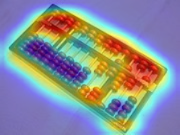}
    \includegraphics[width=1.0\textwidth,height=1.0\textwidth]{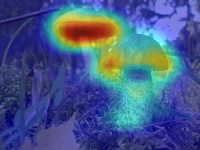}
  \end{minipage}
  \begin{minipage}[t]{0.115\textwidth}
    \setlength{\abovecaptionskip}{0pt}
    \caption*{Fusion}
    \includegraphics[width=1.0\textwidth,height=1.0\textwidth]{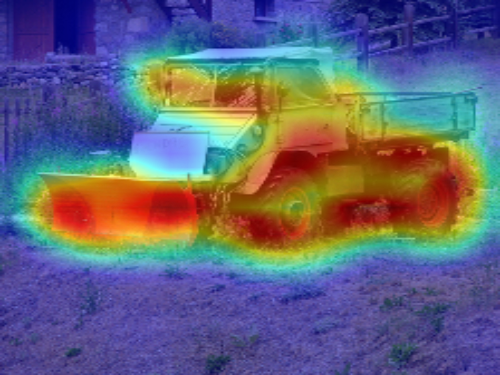}
    \includegraphics[width=1.0\textwidth,height=1.0\textwidth]{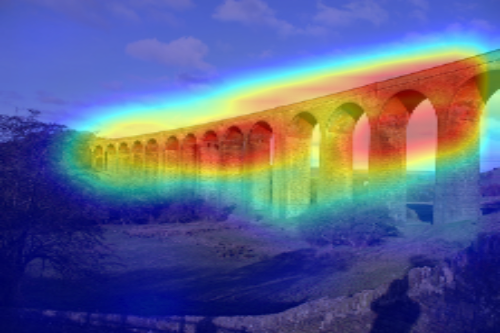}
    \includegraphics[width=1.0\textwidth,height=1.0\textwidth]{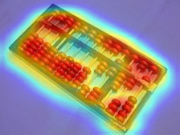}
    \includegraphics[width=1.0\textwidth,height=1.0\textwidth]{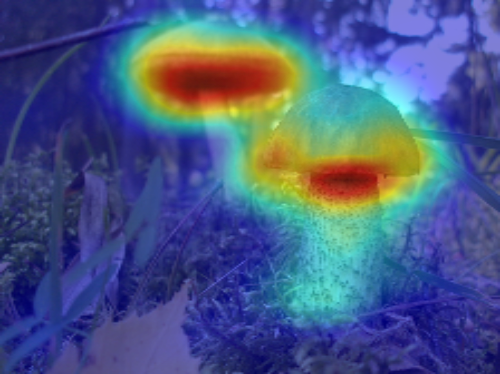}
  \end{minipage}
  }
  \end{subfigure}

  \begin{subfigure}[Caltech256]{
  \begin{minipage}[t]{0.5\textwidth}
  \includegraphics[width=0.23\textwidth,height=0.23\textwidth]{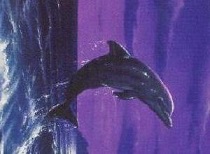}
  \includegraphics[width=0.23\textwidth,height=0.23\textwidth]{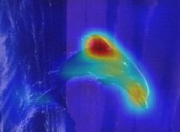}
  \includegraphics[width=0.23\textwidth,height=0.23\textwidth]{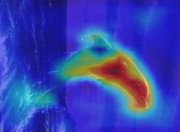}
  \includegraphics[width=0.23\textwidth,height=0.23\textwidth]{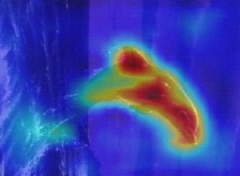}

  \includegraphics[width=0.23\textwidth,height=0.23\textwidth]{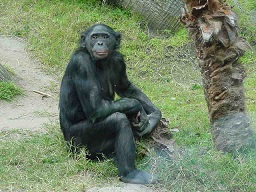}
  \includegraphics[width=0.23\textwidth,height=0.23\textwidth]{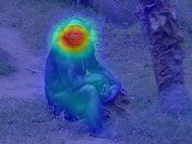}
  \includegraphics[width=0.23\textwidth,height=0.23\textwidth]{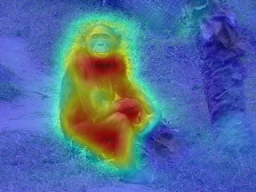}
  \includegraphics[width=0.23\textwidth,height=0.23\textwidth]{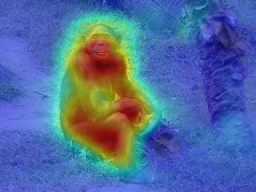}

  \includegraphics[width=0.23\textwidth,height=0.23\textwidth]{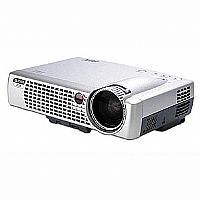}
  \includegraphics[width=0.23\textwidth,height=0.23\textwidth]{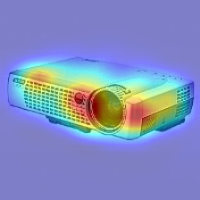}
  \includegraphics[width=0.23\textwidth,height=0.23\textwidth]{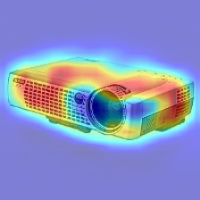}
  \includegraphics[width=0.23\textwidth,height=0.23\textwidth]{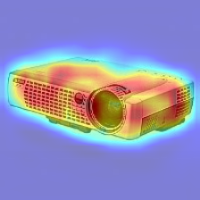}

  \includegraphics[width=0.23\textwidth,height=0.23\textwidth]{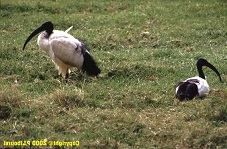}
  \includegraphics[width=0.23\textwidth,height=0.23\textwidth]{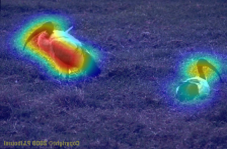}
  \includegraphics[width=0.23\textwidth,height=0.23\textwidth]{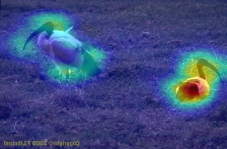}
  \includegraphics[width=0.23\textwidth,height=0.23\textwidth]{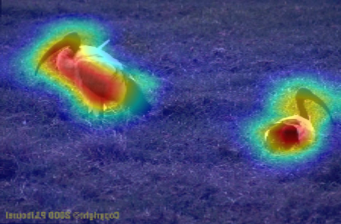}

  \includegraphics[width=0.23\textwidth,height=0.23\textwidth]{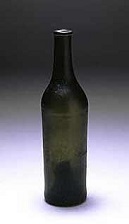}
  \includegraphics[width=0.23\textwidth,height=0.23\textwidth]{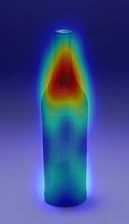}
  \includegraphics[width=0.23\textwidth,height=0.23\textwidth]{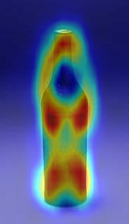}
  \includegraphics[width=0.23\textwidth,height=0.23\textwidth]{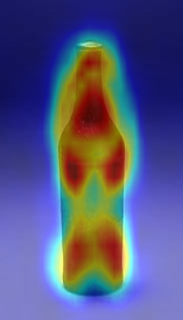}

  \end{minipage}
  }
  \end{subfigure}
  \begin{subfigure}[CUB-200-2011]{
  \begin{minipage}[t]{0.5\textwidth}
  \includegraphics[width=0.23\textwidth,height=0.23\textwidth]{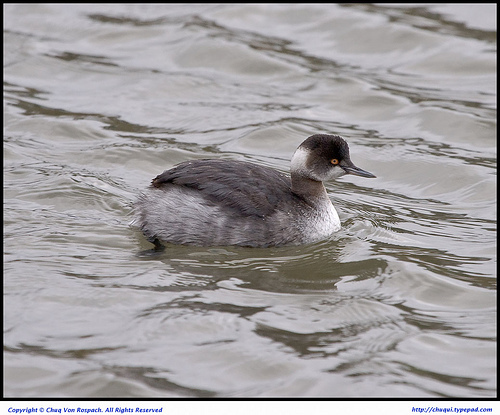}
  \includegraphics[width=0.23\textwidth,height=0.23\textwidth]{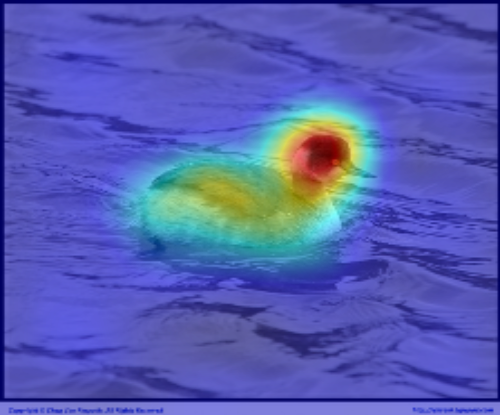}
  \includegraphics[width=0.23\textwidth,height=0.23\textwidth]{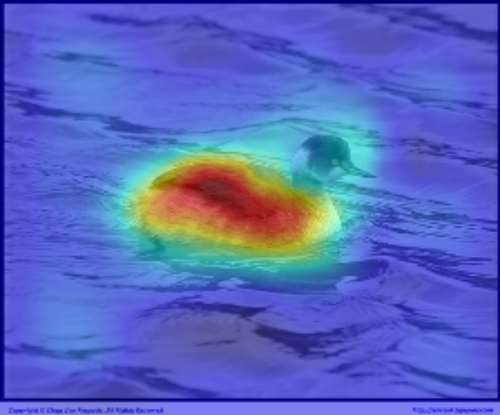}
  \includegraphics[width=0.23\textwidth,height=0.23\textwidth]{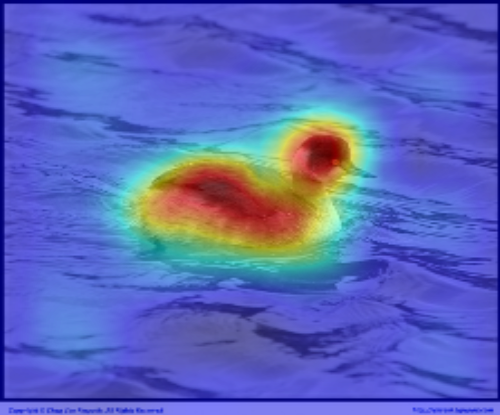}

  \includegraphics[width=0.23\textwidth,height=0.23\textwidth]{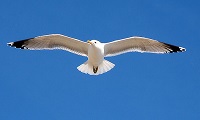}
  \includegraphics[width=0.23\textwidth,height=0.23\textwidth]{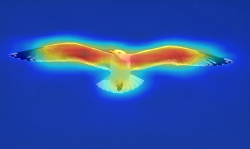}
  \includegraphics[width=0.23\textwidth,height=0.23\textwidth]{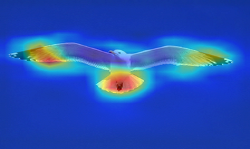}
  \includegraphics[width=0.23\textwidth,height=0.23\textwidth]{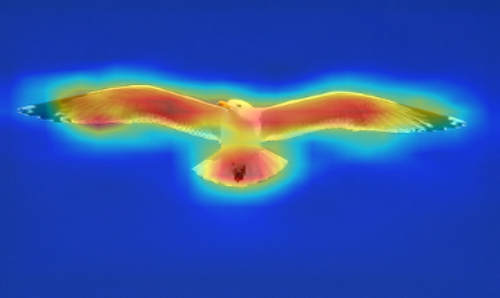}

  \includegraphics[width=0.23\textwidth,height=0.23\textwidth]{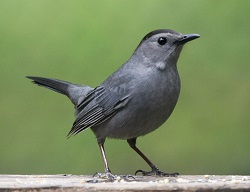}
  \includegraphics[width=0.23\textwidth,height=0.23\textwidth]{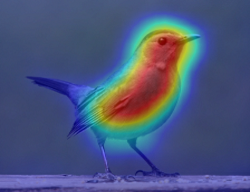}
  \includegraphics[width=0.23\textwidth,height=0.23\textwidth]{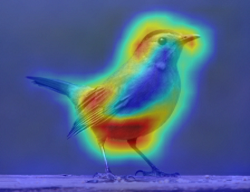}
  \includegraphics[width=0.23\textwidth,height=0.23\textwidth]{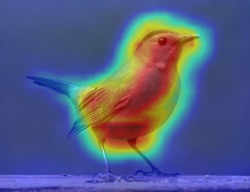}

  \includegraphics[width=0.23\textwidth,height=0.23\textwidth]{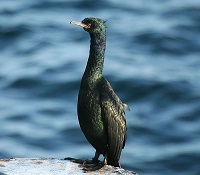}
  \includegraphics[width=0.23\textwidth,height=0.23\textwidth]{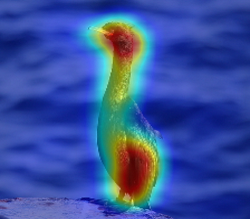}
  \includegraphics[width=0.23\textwidth,height=0.23\textwidth]{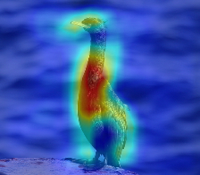}
  \includegraphics[width=0.23\textwidth,height=0.23\textwidth]{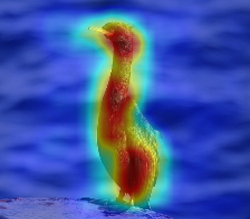}

  \includegraphics[width=0.23\textwidth,height=0.23\textwidth]{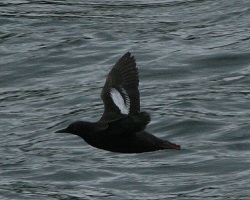}
  \includegraphics[width=0.23\textwidth,height=0.23\textwidth]{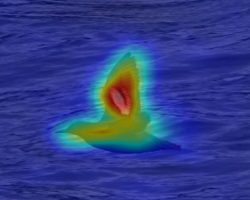}
  \includegraphics[width=0.23\textwidth,height=0.23\textwidth]{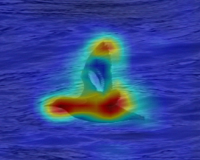}
  \includegraphics[width=0.23\textwidth,height=0.23\textwidth]{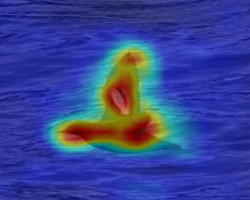}

  \end{minipage}
  }
  \end{subfigure}

  \caption{Object localization maps of the proposed method. We compare complementary effects of the two branches on ILSVRC, Caltech256 and CUB-200-2011 datasets. For each image, we show object localization maps from  Classifier A (middle left), Classifier B (middle right) and the fused maps (right). The proposed two classifier (A and B) can discover different parts of target objects so as to locate the entire regions of the same category in a given image.}\label{fig5}
\end{figure*}

Figure \ref{bbox} visualizes the localization bounding boxes of the proposed method and CAM method~\cite{zhou2015cnnlocalization}.
The object localization maps generated by ACoL can cover larger object regions to obtain more accurate bounding boxes. For example, our method can discover nearly entire parts of a bird, \eg, the wing and head, while the CAM method \cite{zhou2015cnnlocalization} can only find a small part of a bird, \eg, the head.
Figure \ref{fig5} compares the object localization maps of the two classifiers in mining object regions. We observe that Classifier A and Classifier B are successful in discovering different but complementary target regions.
The localization maps from the two classifiers can finally fuse into a robust one, in which the integral object is effectively highlighted. Consequently, we get boosted localization performance.

\subsection{Ablation study}
In the proposed method, the two classifiers locate different regions of interest via erasing the input feature maps of Classifier B. We identify the discriminative regions by a hard threshold $\delta$. In order to inspect its influence on localization accuracy, we test different threshold values $\delta \in \{0.5,0.6,0.7,0.8,0.9\}$ shown in Table \ref{tab6}. We obtain the best performance in Top-1 error when the threshold $\delta=0.6$ on ILSVRC, and it becomes worse when the erasing threshold is larger or smaller. We can conclude: 1) The proposed complementary branch (Classifier B) successfully works collaboratively with Classifier A, because the former can mine complementary object regions so as to generate integral object regions; 2) a well-designed threshold can improve the performance as a too large threshold cannot effectively encourage Classifier B to discover more useful regions and a too small threshold may bring background noises.
\begin{table}\setlength{\tabcolsep}{10pt}
  \centering
  \small
  \begin{tabular}{c|c|c|c}
    \hline
    \hline
    Dataset & threshold & top-1 err. & top-5 err. \\
    \hline
    \multirow{5}{*}{CUB-200-2011} & 0.5 & 58.34 & 48.11 \\
     & 0.6 & 54.15 & \textbf{42.79} \\
     & 0.7 & \textbf{54.08} & 43.49 \\
     & 0.8 & 55.78 &45.17 \\
     & 0.9 & 55.22 &45.76 \\
    \hline
    \multirow{5}{*}{ILSVRC}
    & 0.5 & 62.62 & 52.03 \\
    & 0.6 & \textbf{54.17} & \textbf{40.57} \\
    & 0.7 & 54.55 & 42.53 \\
    & 0.8 & 56.61 & 45.45 \\
    & 0.9 & 55.72 & 44.42 \\
    \hline
    \hline
  \end{tabular}
  \caption{Localization error with different erasing thresholds.}\label{tab6}
  \vspace{-8mm}
\end{table}

%\vspace{-2pt}
We also test a cascade network of three classifiers. In particular, we add the third classifier and erase its input feature maps guided by the fused object localization maps from both Classifier A and B. We observe there is no significant improvement in both classification and localization performance. Therefore, adding the third branch does not necessarily improve the performance and two branches are usually enough for locating the integral object regions.

%\vspace{-2pt}
Furthermore, we eliminate the influence caused by classification results and compare the localization accuracy using ground-truth labels. As shown in Table \ref{tab-gtloc}, the proposed ACoL approach achieves 37.04\% in Top-1 error and surpasses the other approaches. This reveals the superiority of the object localization maps generated by our method, and shows that the proposed two classifiers can successfully locate complementary object regions.
\begin{table}\setlength{\tabcolsep}{10pt}
  \centering
  \small
  \begin{tabular}{l|c}
    \hline
    \hline
    Methods & GT-known loc. err. \\
    \hline
    AlexNet-GAP \cite{zhou2015cnnlocalization} & 45.01 \\
    AlexNet-HaS \cite{singh2017hide} & 41.26 \\
    AlexNet-GAP-ensemble \cite{zhou2015cnnlocalization} & 42.98 \\
    AlexNet-HaS-emsemble \cite{singh2017hide} & 39.67 \\
    GoogLeNet-GAP \cite{zhou2015cnnlocalization} & 41.34 \\
    GoogLeNet-HaS \cite{singh2017hide} & 39.43 \\
    Deconv \cite{zeiler2014visualizing} & 41.6 \\
    Feedback \cite{cao2015look} & 38.8 \\
    MWP \cite{zhang2016top} & 38.7 \\
    \hline
    ACoL (Ours) & \textbf{37.04} \\
    \hline
    \hline
  \end{tabular}
  \caption{Localization error on ILSVRC validation data with ground-truth labels.}\label{tab-gtloc}
  \vspace{-5mm}
\end{table}
%-------------------------------------------------------------------------
%\vspace{-5mm}
\section{Conclusion}
We firstly mathematically prove that object localization maps can be conveniently obtained by selecting from feature maps. Based on it, we proposed Adversarial Complementary Learning for locating target object regions in a weakly supervised manner. The proposed two adversarial classification classifiers can locate different object parts and discover the complementary regions belonging to the same objects or categories. Extensive experiments show the proposed method can successfully mine integral object regions and outperform the state-of-the-art localization methods.
%-------------------------------------------------------------------------
\vspace{-4mm}
\section*{Acknowledgement}
Yi Yang is the recipient of a Google Faculty Research Award.
This work is partially supported by IBM-ILLINOIS Center for Cognitive Computing Systems Research (C3SR) - a research collaboration as part of the IBM AI Horizons Network.
We acknowledge the Data to Decisions CRC (D2D CRC) and the Cooperative Research Centres Programme for funding this research.

{\small
\bibliographystyle{ieee}
\bibliography{egbib}
}

\end{document}